\documentclass[10pt, a4paper]{article}

\let\Item\item
\newcommand\SpecialItem{\renewcommand\item[1][]{\Item[\textbullet~\bfseries##1]}}
\renewcommand\enddescription{\endlist\global\let\item\Item}

\usepackage{lrec2016}
\usepackage{float}
\usepackage{multibib}
\newcites{languageresource}{Language Resources}
\usepackage{graphicx}

\usepackage{amsmath}
\usepackage{epstopdf}
\usepackage[utf8]{inputenc}

\usepackage{url}
\usepackage[normalem]{ulem} 

\usepackage{fancyhdr}
\usepackage{todonotes}
\presetkeys{todonotes}{inline}{}
\presetkeys{todonotes}{prepend}{}
\presetkeys{todonotes}{caption=TODO}{}

\definecolor{darkgreen}{rgb}{0.0, 0.5, 0.0}

\def\ODdel#1{\bgroup\markoverwith{\textcolor{darkgreen}{\rule[0.5ex]{2pt}{1pt}}}\ULon{#1}}

\title{Data Collection for Interactive Learning through the Dialog}

\name{Miroslav Vodolán, Filip Jurčíček}

\address{Charles University in Prague, Faculty of Mathematics and Physics \\
		 Institute of Formal and Applied Linguistics \\
         Malostranské náměstí 25, 11800 Praha 1, Czech Republic\\
         \{vodolan, jurcicek\}@ufal.mff.cuni.cz\\}

\abstract{
This paper presents a dataset collected from natural dialogs which enables to test the ability of dialog systems to learn new facts from user utterances throughout the dialog. 
This \emph{interactive learning} will help with one of the most prevailing problems of open domain dialog system, which is the sparsity of facts a dialog system can reason about. The proposed dataset, consisting of 1900 collected dialogs, allows simulation of an interactive gaining of denotations and questions explanations from users which can be used for the \emph{interactive learning}. 
\\ 
\newline 
}

\begin{document}
\maketitleabstract

\section{Introduction}\label{sec:introduction}
Nowadays, dialog systems are usually designed for a single domain \cite{mrksic_2015}. They store data in a well-defined format with a fixed number of attributes for entities that the system can provide. Because data in this format can be stored as a two-dimensional table within a relational database, we call the data flat. This data representation allows the system to query the database in a simple and efficient way. It also allows to keep the dialog state in the form of slots (which usually correspond to columns in the table) and track it through the dialog using probabilistic belief tracking~\cite{Williams2013,henderson2014second}. 

However, the well-defined structure of the database of a typical dialog system comes with a high cost of extending it as every piece of new information has to fit the format. This is especially a problem when we one is adapting the system for a new domain because its entities could have different attributes. 

A dialog system based on knowledge bases offers many advantages. First, the knowledge base, which can be represented as knowledge graph containing entities connected by relations, is much more flexible than the relational database. Second, freely available knowledge bases, such as Freebase, Wikidata, etc. contain an enormous amount of structured information, and are still growing. A dialog system which is capable of working with this type of information would be therefore very useful. 

In this paper we propose a dataset aiming to help develop and evaluate dialog systems based on knowledge bases by \emph{interactive learning} motivated in Section~\ref{sec:motivation} Section~\ref{sec:dialog_strategies} describes policies that can be used for retrieving information from knowledge bases. In Section~\ref{sec:dialog_simulation} is introduced a dialog simulation from natural conversations which we use for evaluation of \emph{interactive learning}. The dataset collection process allowing the dialog simulation is described in Section~\ref{sec:dataset_collection} and is followed by properties of the resulting dataset in Section~\ref{sec:dataset_splits} Evaluation guidelines with proposed metrics can be found in Section~\ref{sec:interactive_learning_evaluation} The planned future work is summarized in Section~\ref{sec:future_work} We conclude the paper with Section~\ref{sec:conclusion}

\section{Motivation}\label{sec:motivation}
From the point of view of dialog systems providing general information from a knowledge base, the most limiting factor is that a large portion of the questions is understood poorly.

Current approaches~\cite{berant2015imitation,Bordes2014} can only achieve around 50\% accuracy on some question answering datasets. Therefore, we think that there is a room for improvements which can be achieved by interactively asking for additional information in conversational dialogs with users. This extra information can be used for improving policies of dialog systems. We call this approach the \emph{interactive learning} from dialogs. 

We can improve dialog systems in several aspects through \emph{interactive learning} in a direct interaction with users. First, the most straightforward way obviously is getting the correct answer for questions that the system does not know. We can try to ask users for answers on questions that the system encountered in a conversation with a different user and did not understand it. Second, the system can ask the user for a broader explanation of a question. This explanation could help the system to understand the question and provide the correct answer. In addition, the system can learn correct policy for the question which allows providing answers without asking any extra information for similar questions next time. We hypothesize that users are willing to give such explanations because it could help them to find answers for their own questions. The last source of information that we consider for \emph{interactive learning} is rephrasing, which could help when the system does know the concept but does not know the correct wording. This area is extensively studied for the purposes of information retrieval \cite{Imielinski2009,France03learningparaphrases}.

The main purpose of the collected dataset is to enable \emph{interactive learning} using the steps proposed above and potentially to evaluate how different systems perform on this task.

\section{Dialog policies}\label{sec:dialog_strategies}
The obvious difficulty when developing a dialog system is finding a way how to identify the piece of information that the user is interested in. This is especially a problem for dialog systems based on knowledge graphs containing a large amount of complex structured information. While a similar problem is being solved in a task of question answering, dialog systems have more possibilities of identifying the real intention of the user. For example, a dialog system can ask for additional information during the dialog.

We distinguish three different basic approaches to requesting knowledge bases:
\begin{description}
	\item[handcrafted policy] -- the policy consists of fixed set of rules implemented by system developers,

	\item[offline policy] -- the policy is learned from some kind of offline training data (usually annotated) without interaction with system users \cite{Bordes2015}, 

	\item[interactively learned policy] -- the system learns the policy through the dialog from its users by interactively asking them for additional information. 
    

\end{description}

A combination of the above approaches is also possible. For example, we can imagine scenarios where the dialog system starts with hand-crafted rules, which are subsequently interactively improved through dialogs with its users.
With a growing demand for open domain dialog systems, it shows that creating hand-crafted policies does not scale well - therefore, machine learning approaches are gaining on popularity. Many public datasets for offline learning have been published~\cite{Berant2013,Bordes2015}. However, to our knowledge, no public datasets for interactive learning are available.
To fill this gap, we collected a dataset which enables to train interactively learned policies through a simulated interaction with users.

\section{Dialog Simulation}\label{sec:dialog_simulation}

Offline evaluation of interactive dialogs on real data is difficult because different policies can lead to different variants of the dialog. Our solution to this issue is to collect data in a way that allows us to simulate all dialog variants possible according to any policy.

The dialog variants we are considering for \emph{interactive learning} differ only in presence of several parts of the dialog. Therefore, we can collect dialogs containing all information used for interactive learning and omit those parts that were not requested by the policy.

\begin{figure}
    \begin{center}
	    \includegraphics[scale=0.9]{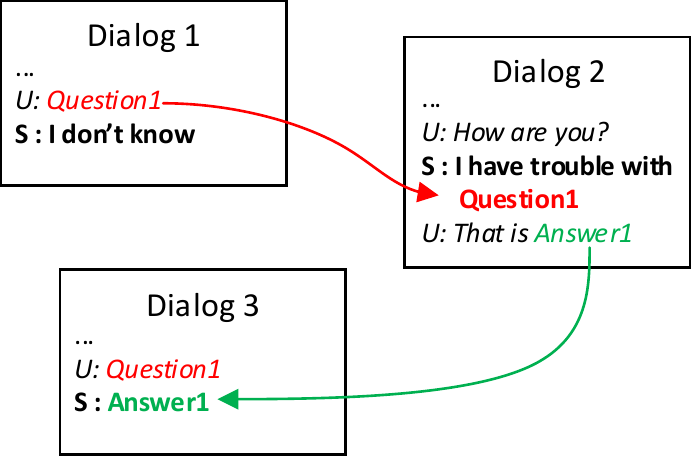}
	\end{center}
	
    \caption{Unknown questions can be rerouted between users. We can, for example, use  chitchat to get correct answers. The challenge is in generalizing the collected question-answer pairs using the knowledge base in order to apply them to previously unseen questions. }
    \label{fig:dialog_system_scheme}
\end{figure}

We collected the dataset (see Section~\ref{sec:dataset_collection}) that enables simulation where the policy can decide how much extra information to the question it requests. If the question is clear to the system it can attempt to answer the question without any other information. It can also ask for a broader explanation with a possibility to answer the question afterwards. If the system decides not to answer the question, we can simulate rerouting the question to another user, to try to obtain the answer from them. The principle of simulated user's answer is shown in the Figure~\ref{fig:dialog_system_scheme}.

Note that the simulated user’s answer can be incorrect because human users naturally made mistakes. We intentionally keep these mistakes in the dataset because real systems must address them as well.

\section{Dataset Collection Process}\label{sec:dataset_collection}
A perfect data collection scenario for our dataset would use real running dialog system providing general information from the knowledge base to real users. This system could then ask for explanations and answers for questions which it is not able to answer.

However, getting access to systems with real users is usually hard. Therefore, we used the crowdsourcing platform CrowdFlower\footnote{\url{http://crowdflower.com}} (CF) for our data collection. 

A CF worker gets a task instructing them to use our chat-like interface to help the system with a question which is randomly selected from training examples of Simple questions~\cite{Bordes2015} dataset. To complete the task user has to communicate with the system through the three phase dialog discussing question paraphrase (see Section~\ref{sec:question_paraphrasing}), explanation (see Section~\ref{sec:question_explanation}) and answer of the question (see Section~\ref{sec:question_answer}). To avoid poor English level of dialogs we involved  CF workers from English speaking countries only. The collected dialogs has been annotated (see Section~\ref{sec:annotation}) by expert annotators afterwards. 

\begin{figure}
    \begin{center}
	    \includegraphics[scale=0.9]{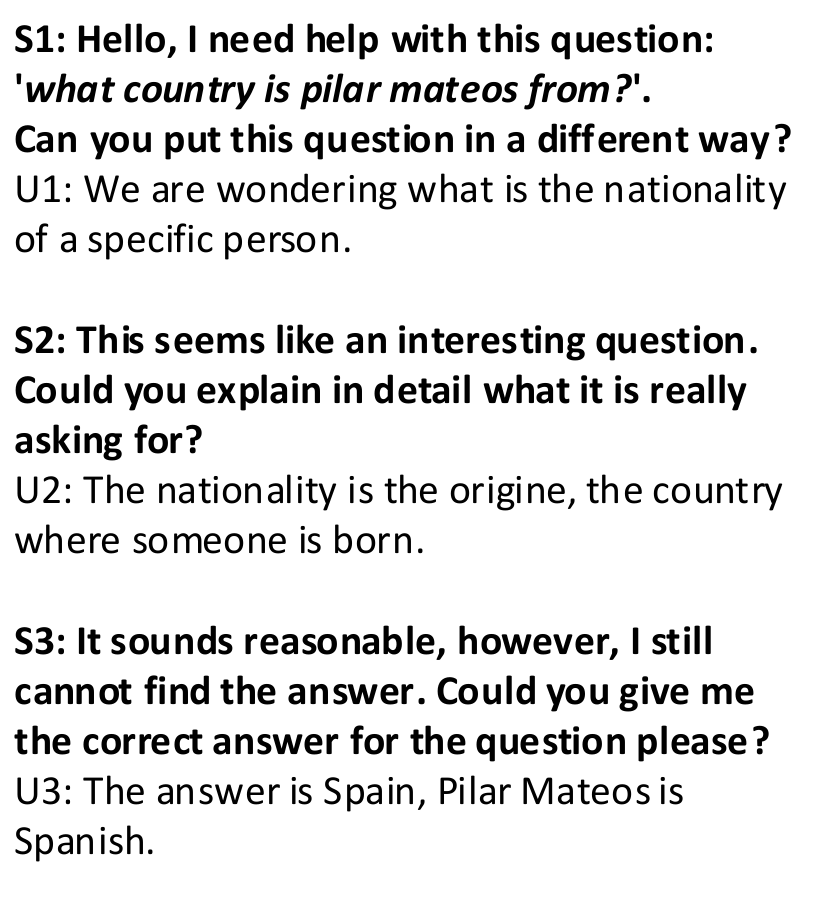}
	\end{center}
	
    \caption{An Example of a short dialog collected on the crowdsourcing platform. We can see that the user provides the question paraphrase (S1), the explanation (S2) and the correct answer for the question (S3).}
    \label{fig:example_dialog}
\end{figure}

The described procedure leads to dialogs like the one shown in the Figure~\ref{fig:example_dialog}.

\subsection{Question Paraphrasing}\label{sec:question_paraphrasing}
At beginning of the dialog, the system is requesting the user to paraphrase question that the system does not understand. The main goal of this first phase is to let the user get familiar with the presented question and to get alternative wordings of the posed question. 
\subsection{Question Explanation}\label{sec:question_explanation}
In the second phase, the user is asked for an explanation of the question. We expect the explanation to be different enough from the original question (in terms of the number of common words between the question and the explanation).
If the explanation is too similar to the question, the user is notified that their explanation is not broad 
enough and they must provide a better one. 
\subsection{Question Answer}\label{sec:question_answer}
With the valid explanation the dialog turns into the last phase where the user is asked for a correct answer to the original question. 
The system requires the user to answer with a full sentence. In practical experiments this has shown as a useful decision because it improves system's ability to reveal cheaters. We can simply measure the connection (in terms of common words 
) between question and the answer sentence. This allows to reject completely irrelevant answers.
\subsection{Annotation}\label{sec:annotation}
The correct answer for question in each dialog is available from Simple questions dataset.
Answers are in form of Freebase\footnote{\url{https://www.freebase.com/}} entities identified by unique id.
For evaluation purposes we need information whether dialog contains the answer which is consistent with the entity from Simple questions, the answer with another entity or whether the dialog does not contain any answer. While the annotation process is quite simple, we did not need crowdsourcing for the process.

\subsection{Natural Language Understanding (NLU)}\label{sec:nlu}
The collection system needs to recognize following dialog acts from user utterances during all phases of the dialog:
\begin{description}
	\item[Negate] -- user does not want to provide requested information,
	\item[Affirm] -- user agrees to provide requested information,
    \item[DontKnow] -- user does not know the requested information,
    \item[ChitChat] -- user tries chit chat with the system (hello, bye, who are you...),
    \item[Inform] -- none of the above, interpreted as user is giving information requested by the system.
\end{description}

Parsing of the dialog acts is made by hand written rules using templates and keyword spotting. The templates and keywords were manually collected from frequent expressions used by CF workers during preparation runs of the dataset collection process (google it, check wikipedia, I would need...~$\rightarrow$~Negate).

\section{Dataset Properties}\label{sec:dataset_splits}
We collected the dataset with 1900 dialogs and 8533 turns. 
Topics discussed in dialogs are questions randomly chosen from training examples of Simple questions~\cite{Bordes2015} dataset. From this dataset we also took the correct answers in form of Freebase entities.

Our dataset consists of standard data split into training, development and test files. The basic properties of those files are as follows:

\begin{table}[H]
\begin{center}
\bgroup
\def\arraystretch{1.05}
\setlength{\tabcolsep}{3pt}
\begin{tabular}{ll||c|c}
& & dialog count & dialog turns\\[0.1cm] \hline
 
& Training dialogs &  950 & 4249\\ 
& Development dialogs &  285 &  1258\\ \hline
& Testing dialogs &  665 & 3026\\ \hline

\end{tabular}
\egroup
\end{center}
\caption{Table of turn and dialog counts for dataset splits.}
\label{tab:split_statistics}
\end{table}

Each file contains complete dialogs enriched by outputs of NLU (see Section~\ref{sec:nlu}) that were used during the data collection. On top of that, each dialog is labeled by the correct answer for the question and expert annotation of the user answer hint which tells whether the hint points to the correct answer, incorrect answer, or no answer at all.

351 of all collected dialogs contain correct answer provided by users and 702 dialogs have incorrect answer. In the remaining 847 dialogs users did not want to answer the question. The collected dialogs also contain 1828 paraphrases and 1539 explanations for 1870 questions.

An answer for a question was labeled as correct by annotators only when it was evident to them that the answer points to the same Freebase entity that was present in Simple questions dataset for that particular question.
However, a large amount of questions from that dataset is quite general - with many possible answers. Therefore lot of answers from users were labeled as incorrect even though those answers perfectly fit the question. Our annotators identified that 285 of the incorrect answers were answers for such general questions.
Example of this situation can be demonstrated by question \textit{'Name an actor'} which was correctly answered by \textit{'Brad Pitt is an actor'}, however, to be consistent with Simple questions annotation, which is \textit{'Kelly Atwood'}, annotators were forced to mark it as an incorrect answer.

\section{Interactive Learning Evaluation}\label{sec:interactive_learning_evaluation}
A perfect \emph{interactive learning} model would be able to learn anything interactively from test dialogs during testing, which would allow us to measure progress of the model from scratch over the course of time.
However, a development of such model would be unnecessarily hard, therefore we provide training dialogs which can be used for feature extraction and other engineering related to \emph{interactive learning} from dialogs in natural language. Model development is further supported with labeled validation data for parameter tuning.

We propose two evaluation metrics for comparing \emph{interactive learning} models. First metric (see Section~\ref{sec:efficiency_score}) scores amount of information required by the model, second metric (see Section~\ref{sec:answer_extraction_accuracy}) is accuracy of  answer extraction from user utterances.
All models must base their answers only on information gained from training dialogs and testing dialogs seen during the simulation so far, to ensure that the score will reflect the \emph{interactive learning} of the model instead of general question answering.

\subsection{Efficiency Score}\label{sec:efficiency_score}
The simulation of dialogs from our dataset allows to evaluate how efficient a dialog system is in using information gained from users. The dialog system should maximize the number of correctly answered questions without requesting too many explanations and answers from users. To evaluate different systems using the collected data, we propose the following evaluation measure:

\begin{equation} \label{eq:score}
	S_D = \frac{n_c - w_i n_i - w_e n_e - w_a n_a}{|D|}
\end{equation}

Here, $n_c$ denotes the number of correctly answered questions, $n_i$ denotes the number of incorrectly answered questions, $n_e$ denotes the number of requested explanations, $n_a$ denotes the number of requested answers and $|D|$ denotes the number of simulated dialogs in the dataset. $w_i$, $w_e$, $w_a$ are penalization weights. 

The penalization weights are used to compensate for different costs of obtaining different types of information from the user. For example, gaining broader explanation from the user is relatively simple because it is in their favor to cooperate with the system on a question they are interested in. However, obtaining correct answers from users is significantly more difficult because the system does not always have the chance to ask the question and the user does not have to know the correct answer for it.

To make the evaluations comparable between different systems we recommend using our evaluation scripts included with the dataset with following penalization weights that reflect our intuition for gaining information from users: 

\SpecialItem
\begin{description}
  \item	[$w_i = 5$] – incorrect answers are penalized significantly,
  \item [$w_e = 0.2$] – explanations are quite cheap; therefore, we will penalize them just slightly,
  \item [$w_a = 1$] – gaining question’s answer from users is harder than gaining explanations.
\end{description}

\subsection{Answer Extraction Accuracy}\label{sec:answer_extraction_accuracy}
It is quite challenging to find appropriate entity in the knowledge base even though the user provided the correct answer. Therefore, we propose another metric relevant to our dataset. This metric is the accuracy of entity extraction which measures how many times was extracted a correct answer from  answer hints provided by the user in dialogs annotated as correctly answered. 

\section{Future Work}\label{sec:future_work}
Our future work will be mainly focused on providing a baseline system for interactive learning which will be evaluated on the dataset. We are also planning improvements for dialog management that is used to gain explanations during the data collection. We believe that with conversation about specific aspects of the discussed question it will be possible to gain even more interesting information from users.
The other area of our interest is in possibilities to improve question answering accuracy on test questions of Simple question dataset with the extra information contained in the collected dialogs.

\section{Conclusion}\label{sec:conclusion}
In this paper, we presented a novel way how to evaluate different interactive learning approaches for dialog models. The evaluation covers two challenging aspects of interactive learning. First, it scores efficiency of using information gained from users in simulated question answering dialogs. Second, it measures accuracy on answer hints understanding. 

For purposes of evaluation we collected a dataset from conversational dialogs with workers on crowdsourcing platform CrowdFlower. Those dialogs were annotated with expert annotators and published under Creative Commons~4.0 BY-SA license on lindat\footnote{\url{hdl.handle.net/11234/1-1670}}. We also provide evaluation scripts with the dataset that should ensure comparable evaluation of different interactive learning approaches.

\section{Acknowledgments}
This  work  was  funded  by  the  Ministry  of  Education,  Youth  and  Sports  of  the  Czech  Republic
under the grant agreement LK11221 and core research funding, SVV project 260 224, and GAUK
grant 1170516 of Charles University in Prague. It used language resources stored and distributed by
the LINDAT/CLARIN project of the Ministry of Education, Youth and Sports of the Czech Republic (project  LM2015071).

\section{Bibliographical References}
\label{main:ref}

\bibliographystyle{lrec2016}
\bibliography{paper}

\end{document}